\documentclass[10pt,twocolumn,letterpaper]{article}

\usepackage{cvpr}              

\usepackage[dvipsnames]{xcolor}

\definecolor{cvprblue}{rgb}{0.21,0.49,0.74}
\usepackage[pagebackref,breaklinks,colorlinks,citecolor=cvprblue]{hyperref}

\usepackage{hyperref}
\usepackage{url}
\usepackage{graphicx}

\usepackage{booktabs}
\usepackage{tablefootnote}
\usepackage[flushleft]{threeparttable}

\usepackage{lipsum}
\usepackage{bbding}
\usepackage{url}

\usepackage{color}
\usepackage{enumitem}

\usepackage{multirow}
\usepackage{ulem}

\usepackage{url}            
\usepackage{tcolorbox}       
\usepackage{amsfonts}       
\usepackage{nicefrac}       
\usepackage{microtype}      
\usepackage{xcolor}
\definecolor{mygreen}{HTML}{3cb44b}

\usepackage[accsupp]{axessibility}
\def\paperID{8927} 
\def\confName{CVPR}
\def\confYear{2024}

\title{LISA: Reasoning Segmentation via Large Language Model}

\author{Xin Lai$^{1}$\thanks{Equal Contribution}\hspace{0.7cm}Zhuotao Tian$^{2*}$\thanks{Corresponding Author (\href{ tianzhuotao@hit.edu.cn}{ tianzhuotao@hit.edu.cn}).}\hspace{0.5cm}Yukang Chen$^{1}$\hspace{0.5cm}Yanwei Li$^{1}$\hspace{0.5cm}Yuhui Yuan$^{4}$\hspace{0.5cm}Shu Liu$^{3}$\hspace{0.5cm}Jiaya Jia$^{1,3}$\\
$^{1}$CUHK~~~
$^{2}$HIT (Shenzhen)~~~
$^{3}$SmartMore~~~
$^{4}$MSRA~~~
}

\begin{document}
\maketitle
\begin{figure*}[h]
\vspace{-0.3cm}
\begin{center}
\includegraphics[width=1.0\linewidth]{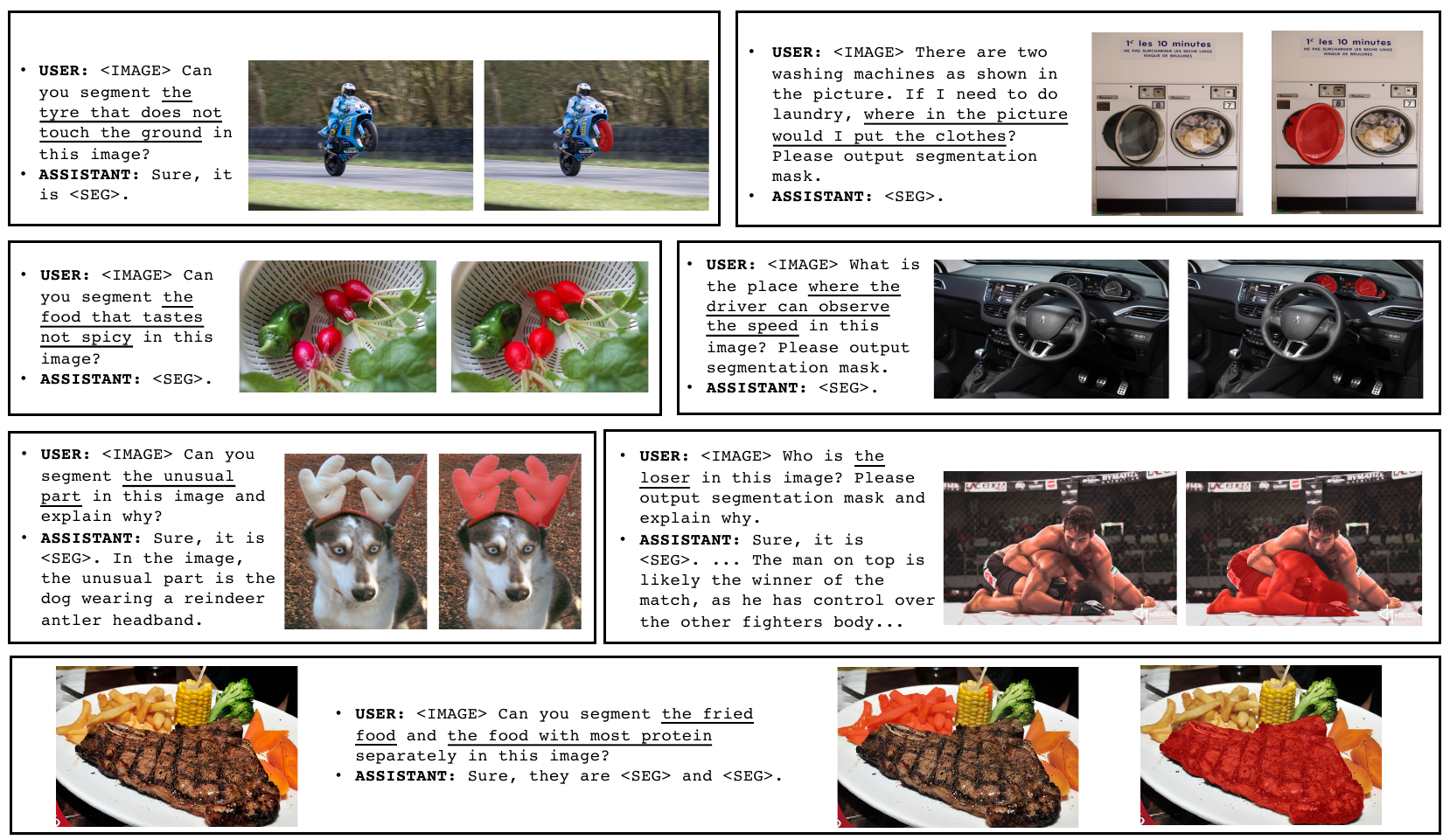}
\end{center}
\vspace{-0.6cm}
\caption{We unlock new segmentation capabilities for existing multimodal LLMs. Our model (i.e., LISA) can deal with cases involving complex reasoning and world knowledge. Also, we demonstrate the cases of explanatory answers in the 3rd row. Additionally, in the 4th row, our model can output multiple segmentation masks in a single answer. More illustrations can be found in the supplementary material.}
\label{fig:teaser}
\vspace{-0.2cm}
\end{figure*}

\begin{abstract}
Although perception systems have made remarkable advancements in recent years, they still rely on explicit human instruction or pre-defined categories to identify the target objects before executing visual recognition tasks. Such systems cannot actively reason and comprehend implicit user intention. In this work, we propose a new segmentation task --- \textit{reasoning segmentation}. The task is designed to output a segmentation mask given a complex and implicit query text. Furthermore, we establish a benchmark comprising over one thousand image-instruction-mask data samples, incorporating intricate reasoning and world knowledge for evaluation purposes. Finally, we present LISA: large \textbf{L}anguage \textbf{I}nstructed \textbf{S}egmentation \textbf{A}ssistant, which inherits the language generation capabilities of multimodal Large Language Models (LLMs) while also possessing the ability to produce segmentation masks. We expand the original vocabulary with a \texttt{<SEG>} token and propose the embedding-as-mask paradigm to unlock the segmentation capability. Remarkably, LISA can handle cases involving complex reasoning and world knowledge. Also, it demonstrates robust zero-shot capability when trained exclusively on reasoning-free datasets. In addition, fine-tuning the model with merely 239 reasoning segmentation data samples results in further performance enhancement. Both quantitative and qualitative experiments show our method effectively unlocks new reasoning segmentation capabilities for multimodal LLMs. Code, models, and data are available at \href{https://github.com/dvlab-research/LISA}{github.com/dvlab-research/LISA}. 
\end{abstract}    
\section{Introduction}
\label{sec:intro}



In daily life, users tend to issue direct commands like ``Change the TV channel" to instruct a robot, rather than providing explicit step-by-step instructions such as ``Go to the table first, find the TV remote, and then press the button to change the channel." However, existing perception systems consistently rely on humans to explicitly indicate target objects or pre-define categories before executing visual recognition tasks. These systems cannot actively reason and comprehend user intention based on implicit instruction. This reasoning ability is crucial in developing next-generation intelligent perception systems and holds substantial potential for industrial applications, particularly in robotics.

In this work, we introduce a new segmentation task --- \textit{reasoning segmentation}, which requires generating a binary segmentation mask based on an implicit query text involving \textit{complex reasoning}. Notably, the query text is not limited to a straightforward reference (e.g., ``the orange"), but a more complicated description involving \textit{complex reasoning} or \textit{world knowledge} (e.g., ``the food with high Vitamin C"). To accomplish this task, the model must possess two key abilities: 1) reasoning \textit{complex} and \textit{implicit} text queries jointly with the image; 2) producing segmentation masks.

Inspired by the exceptional capacity of LLMs to reason and comprehend user intentions, we aim to leverage this capability of LLMs to address the aforementioned first challenge. However, while several studies~\cite{alayrac2022flamingo,li2023blip,ye2023mplug,li2023otter,liu2023visual,zhu2023minigpt,liu2023improved} have integrated robust reasoning capabilities into multimodal LLMs to accommodate visual input, the majority of these models primarily concentrate on text generation tasks and still fall short in performing vision tasks that require fine-grained output formats, such as segmentation masks. This leads us to ask: can we enable multimodal LLMs with the capability to output segmentation masks?


To this end, we introduce LISA: a large \textbf{L}anguage \textbf{I}nstructed \textbf{S}egmentation \textbf{A}ssistant, a multimodal LLM capable of producing segmentation masks. Specifically, we incorporate an additional token, i.e., \texttt{<SEG>}, into the existing vocabulary. Upon generating the \texttt{<SEG>} token, its hidden embedding is further decoded into the corresponding segmentation mask. By representing the segmentation mask as an embedding, LISA acquires segmentation capabilities and benefits from end-to-end training. Remarkably, LISA demonstrates robust zero-shot abilities. Training the model solely on standard semantic segmentation and referring segmentation datasets yields surprisingly effective performance on the reasoning segmentation task. Furthermore, we find that LISA's performance can be significantly enhanced by fine-tuning on just 239 reasoning segmentation data samples. As illustrated in Fig.~\ref{fig:teaser}, LISA can handle various scenarios involving complex reasoning and world knowledge.


In addition, to validate the effectiveness, we establish a benchmark for reasoning segmentation evaluation, called \textit{ReasonSeg}. Comprising over one thousand image-instruction pairs, this benchmark offers persuasive evaluation metrics for the task. To align more closely with practical applications, we annotate the images from OpenImages~\citep{OpenImages} and ScanNetv2~\citep{dai2017scannet} with implicit text queries that involve complex reasoning. 

In summary, our contributions are as follows:
\begin{itemize}[leftmargin=0.7cm]
    \item We introduce the \textit{reasoning segmentation} task, which necessitates reasoning based on implicit human instructions. Such reasoning capability is crucial for building a genuinely intelligent perception system.
    
    

    \vspace{0.2cm}
    \item We present our model --- LISA, which incorporates new segmentation capabilities. It demonstrates robust zero-shot ability on the reasoning segmentation task when trained solely on reasoning-free datasets, and achieves further performance boost by fine-tuning on just 239 data samples that involve reasoning. 
    
    \vspace{0.2cm}
    \item We establish a reasoning segmentation benchmark, \textit{ReasonSeg}, containing over one thousand image-instruction-mask data samples. This benchmark is essential for evaluation and encourages the community to further explore the reasoning ability for vision tasks.
\end{itemize}

\begin{figure*}[t]
\begin{center}
\includegraphics[width=0.88\linewidth]{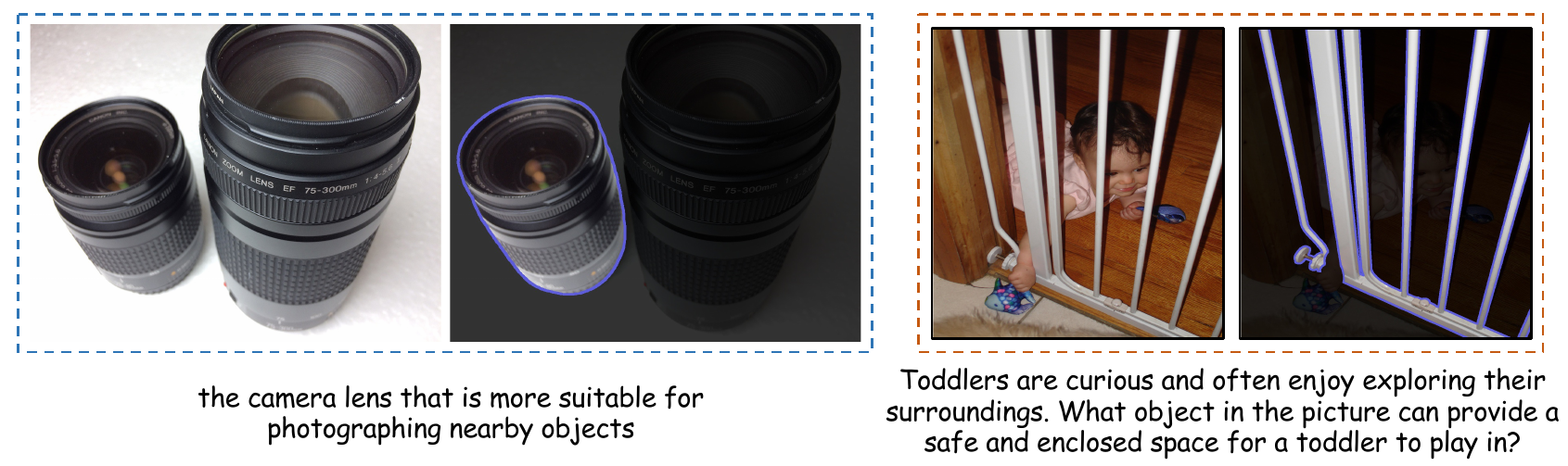}
\end{center}
\vspace{-0.6cm}
\caption{Examples of the annotated image-instruction-mask data samples. Left: short phrase query. Right: long sentence query. More examples are given in the supplementary material.}
\label{fig:benchmark}
\vspace{-0.4cm}
\end{figure*}

\section{Related Work}
\vspace{-0.2cm}
\subsection{Image Segmentation}
\vspace{-0.2cm}

Semantic segmentation aims to assign a class label to every pixel in an image. Numerous studies~\citep{fcn,deconvnet,segnet,unet,deeplab,dilation,parsenet,pspnet,icnet,denseaspp,danet,ccnet,psanet,asymmetric_nonlocal,cheng2021per,lai2021semi,tian2022adaptive,tian2023learning} have proposed diverse designs (such as encoder-decoder, dilated convolution, pyramid pooling module, non-local operator, and more) to effectively encode semantic information. Research on instance segmentation~\citep{he2017mask,zhang2021k,cheng2022masked} and panoptic segmentation~\citep{kirillov2019panoptic,xiong2019upsnet,cheng2020panoptic,li2021fully} has introduced various architectural innovations for instance-level segmentation, including DETR~\citep{carion2020end}-based structures, mask attention, and dynamic convolution. In recent years, typical segmentation tasks have made significant progress and become increasingly mature. Consequently, it is imperative to develop more intelligent interaction ways for image segmentation.


The referring segmentation task~\citep{kazemzadeh2014referitgame,nagaraja2016modeling} enables interaction with human language, aiming to segment the target object based on a given explicit text description. Recently, \citet{kirillov2023segment} introduced SAM, trained with billions of high-quality masks, supporting bounding boxes and points as prompts while demonstrating exceptional segmentation quality. X-Decoder~\citep{zou2023generalized} bridges vision and language, unifying multiple tasks within a single model. SEEM~\citep{zou2023segment} further supports various human interaction methods, including text, audio, and scribble. However, these studies primarily focus on addressing multi-task compatibility and unification, neglecting the injection of new capabilities. In this work, we present LISA and it possesses reasoning ability that has not been explored yet in existing segmentors.

\vspace{-0.1cm}
\subsection{Multimodal Large Language Model}
\vspace{-0.1cm}

Motivated by the remarkable reasoning abilities of LLMs, researchers are exploring ways to transfer these capabilities into the vision domain, developing multimodal LLMs. Flamingo~\citep{alayrac2022flamingo} employs a cross-attention structure to attend to visual contexts, enabling visual in-context learning. Models such as BLIP-2~\citep{li2023blip} and mPLUG-OWL~\citep{ye2023mplug} propose encoding image features with a visual encoder, which are then fed into the LLM alongside text embeddings. Otter~\citep{li2023otter} further incorporates robust few-shot capabilities through in-context instruction tuning on the proposed MIMIC-IT dataset. LLaVA~\citep{liu2023visual} and MiniGPT-4~\citep{zhu2023minigpt} first conduct image-text feature alignment followed by instruction tuning. \citet{koh2023grounding} also investigates image retrieval for LLMs. Moreover, numerous works~\citep{wu2023visual,yang2023mm,shen2023hugginggpt,liu2023internchat,yang2023gpt4tools} utilize prompt engineering, connecting independent modules via API calls, but without the benefits of end-to-end training. Recently, there have been studies examining the intersection between multimodal LLMs and vision tasks. VisionLLM~\citep{wang2023visionllm} offers a flexible interaction interface for multiple vision-centric tasks through instruction tuning but fails to fully exploit LLMs for complex reasoning. Kosmos-2~\citep{peng2023kosmos} constructs large-scale data of grounded image-text pairs, infusing grounding capabilities into LLMs. DetGPT~\citep{detgpt} bridges the fixed multimodal LLM and open-vocabulary detector, enabling detection to be performed based on user instruction.
GPT4RoI~\citep{zhang2023gpt4roi} introduces spatial boxes as input and trains the model on region-text pairs. In contrast, our work aims to efficiently inject segmentation capabilities into multimodal LLMs in the manner of end-to-end training.

\section{Reasoning Segmentation }
\vspace{-0.1cm}
\subsection{Problem Definition}
\vspace{-0.1cm}
The reasoning segmentation task is to output a binary segmentation mask $\mathbf{M}$, given an input image $\mathbf{x}_{img}$ and an implicit query text instruction $\mathbf{x}_{txt}$. The task shares a similar formulation with the referring segmentation task~\citep{kazemzadeh2014referitgame}, but is far more challenging. The key distinction lies in the complexity of the query text in reasoning segmentation. Instead of a straightforward phrase (e.g., ``the trash can"), the query text includes more intricate expressions (e.g., ``something that the garbage should be put into") or longer sentences (e.g., ``After cooking, consuming food, and preparing for food, where can we throw away the rest of the food and scraps?") that involve complex reasoning or world knowledge.


\vspace{-0.05cm}
\subsection{Benchmark}
\vspace{-0.05cm}



Given the lack of quantitative evaluation, it is imperative to establish a benchmark for the reasoning segmentation task. To ensure reliable assessment, we have collected a diverse set of images from OpenImages~\citep{OpenImages} and ScanNetv2~\citep{dai2017scannet}, annotating them with implicit text instructions and high-quality target masks. To cover different scenarios, our text instructions consist of two types: 1) short phrases; 2) long sentences; as illustrated in Figure~\ref{fig:benchmark}. The resulting \textit{ReasonSeg} benchmark comprises a total of 1218 image-instruction-mask data samples. This dataset is further partitioned into three splits: \texttt{train}, \texttt{val}, and \texttt{test}, containing 239, 200, and 779 data samples, respectively. As the primary purpose of the benchmark is evaluation, the validation and testing sets include a larger number of data samples. The details of data annotation are given in the supplementary material.

\begin{figure*}[t]
\begin{center}
\includegraphics[width=0.98\linewidth]{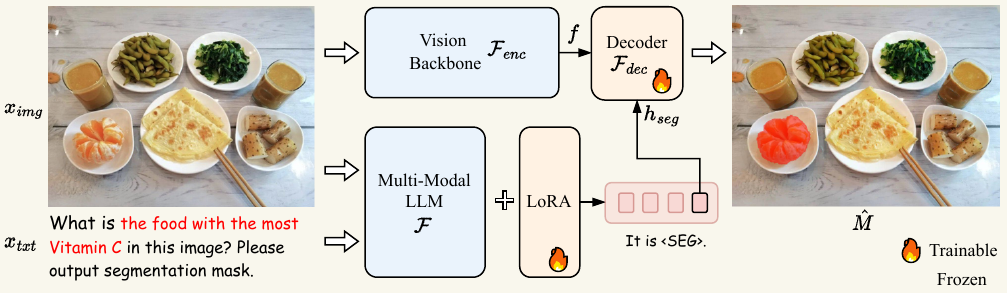}
\end{center}
\vspace{-0.5cm}
\caption{The pipeline of LISA. Given the input image and text query, the multimodal LLM (e.g., LLaVA~\cite{liu2023visual}) generates text output. The last-layer embedding for the \texttt{<SEG>} token is then decoded into the segmentation mask via the decoder. We use LoRA~\cite{hu2021lora} for efficient fine-tuning. The choice of vision backbone can be flexible (e.g., SAM~\cite{zou2023segment}, Mask2Former~\cite{cheng2022masked}).}
\label{fig:overview}
\vspace{-0.4cm}
\end{figure*}

\section{Our Method}

In this section, we first introduce the model architecture in Sec.~\ref{method:arch}. After that, we elaborate on the training data preparation and training parameters in Sec.~\ref{method:training}. 
\subsection{Architecture}
\label{method:arch}


\paragraph{Embedding as Mask. } Most current multimodal LLMs (such as LLaVA~\citep{liu2023visual}, Flamingo~\citep{alayrac2022flamingo}, BLIP-2~\citep{li2023blip}, Otter~\citep{li2023otter}, etc.) support image and text as input, but they can only output text and cannot directly output fine-grained segmentation masks. VisionLLM~\citep{wang2023visionllm} offers a solution by parsing segmentation masks as sequences of polygons, enabling the representation of segmentation masks as plain text and allowing end-to-end training within the framework of existing multimodal LLMs. However, end-to-end training with the polygon sequences introduces optimization challenges and may compromise generalization ability unless a massive amount of data and computational resources are employed. For instance, training a 7B model, VisionLLM requires $4\times 8$ NVIDIA 80G A100 GPUs and 50 epochs, which is computationally prohibitive. In contrast, it takes less than 3 days to train LISA-7B on 8 NVIDIA 24G 3090 GPUs.

To this end, we propose the embedding-as-mask paradigm to infuse new segmentation capabilities into the multimodal LLM. The pipeline of our method is illustrated in Fig.~\ref{fig:overview}. Specifically, we first expand the original LLM vocabulary with a new token, i.e., \texttt{<SEG>}, which signifies the request for the segmentation output. Given a text instruction $\mathbf{x}_{txt}$ along with the input image $\mathbf{x}_{img}$, we feed them into the multimodal LLM $\mathcal{F}$, which in turn outputs a text response $\hat{\mathbf{y}}_{txt}$. It can be formulated as
\begin{align}
\begin{aligned}
    \hat{\mathbf{y}}_{txt} = & \;  \mathcal{F}(\mathbf{x}_{img}, \mathbf{x}_{txt}).
\end{aligned}
\end{align}

When the LLM intends to generate a binary segmentation mask, the output $\hat{\mathbf{y}}_{txt}$ would include a \texttt{<SEG>} token. We then extract the LLM last-layer embedding $\tilde{\mathbf{h}}_{seg}$ corresponding to the \texttt{<SEG>} token and apply an MLP projection layer $\gamma$ to obtain $\mathbf{h}_{seg}$. Simultaneously, the vision backbone $\mathcal{F}_{enc}$ extracts the dense visual features $\mathbf{f}$ from the visual input $\mathbf{x}_{img}$. Finally, $\mathbf{h}_{seg}$ and $\mathbf{f}$ are fed to the decoder $\mathcal{F}_{dec}$ to produce the final segmentation mask $\hat{\mathbf{M}}$. The detailed structure of the decoder $\mathcal{F}_{dec}$ follows \cite{kirillov2023segment}. The process can be formulated as 
\begin{align}
\begin{aligned}
    \mathbf{h}_{seg} = \gamma(\tilde{\mathbf{h}}_{seg} &), \quad \mathbf{f} = \mathcal{F}_{enc}(\mathbf{x}_{img}), \\
    \hat{\mathbf{M}} = & \; \mathcal{F}_{dec}(\mathbf{h}_{seg}, \mathbf{f}).
\end{aligned}
\end{align}



\paragraph{Training Objectives. } The model is trained end-to-end using the text generation loss $\mathcal{L}_{txt}$ and the segmentation mask loss $\mathcal{L}_{mask}$. The overall objective $\mathcal{L}$ is the weighted sum of these losses, determined by $\lambda_{txt}$ and $\lambda_{mask}$:
\begin{equation}
    \mathcal{L} = \lambda_{txt} \mathcal{L}_{txt} + \lambda_{mask} \mathcal{L}_{mask}.
\end{equation}
Specifically, $\mathcal{L}_{txt}$ is the auto-regressive cross-entropy loss for text generation, and $\mathcal{L}_{mask}$ is the mask loss, which encourages the model to produce high-quality segmentation results. To compute $\mathcal{L}_{mask}$, we employ a combination of per-pixel binary cross-entropy (BCE) loss and DICE loss, with corresponding loss weights $\lambda_{bce}$ and $\lambda_{dice}$. Given the ground-truth targets $\mathbf{y}_{txt}$ and $\mathbf{M}$, these losses can be formulated as
\begin{align}
\begin{aligned}
    \mathcal{L}_{txt} & = \mathbf{CE}(\hat{\mathbf{y}}_{txt}, \mathbf{y}_{txt}), \\
    \mathcal{L}_{mask} = \lambda_{bce} \mathbf{BCE}&(\hat{\mathbf{M}}, \mathbf{M})  + \lambda_{dice}\mathbf{DICE}(\hat{\mathbf{M}}, \mathbf{M}).
\end{aligned}
\end{align}

It is noteworthy that the proposed method endows existing multimodal LLMs with new segmentation capabilities, such that they can generate not only text but also fine-grained output formats. Also, our method is based on an end-to-end training pipeline and connects the LLM and vision modules with hidden embedding representation, which proves significantly more effective than the decoupled two-stage method as discussed in Sec.~\ref{exp:reasonseg}. 

\begin{figure*}[t]
\begin{center}
\includegraphics[width=1.0\linewidth]{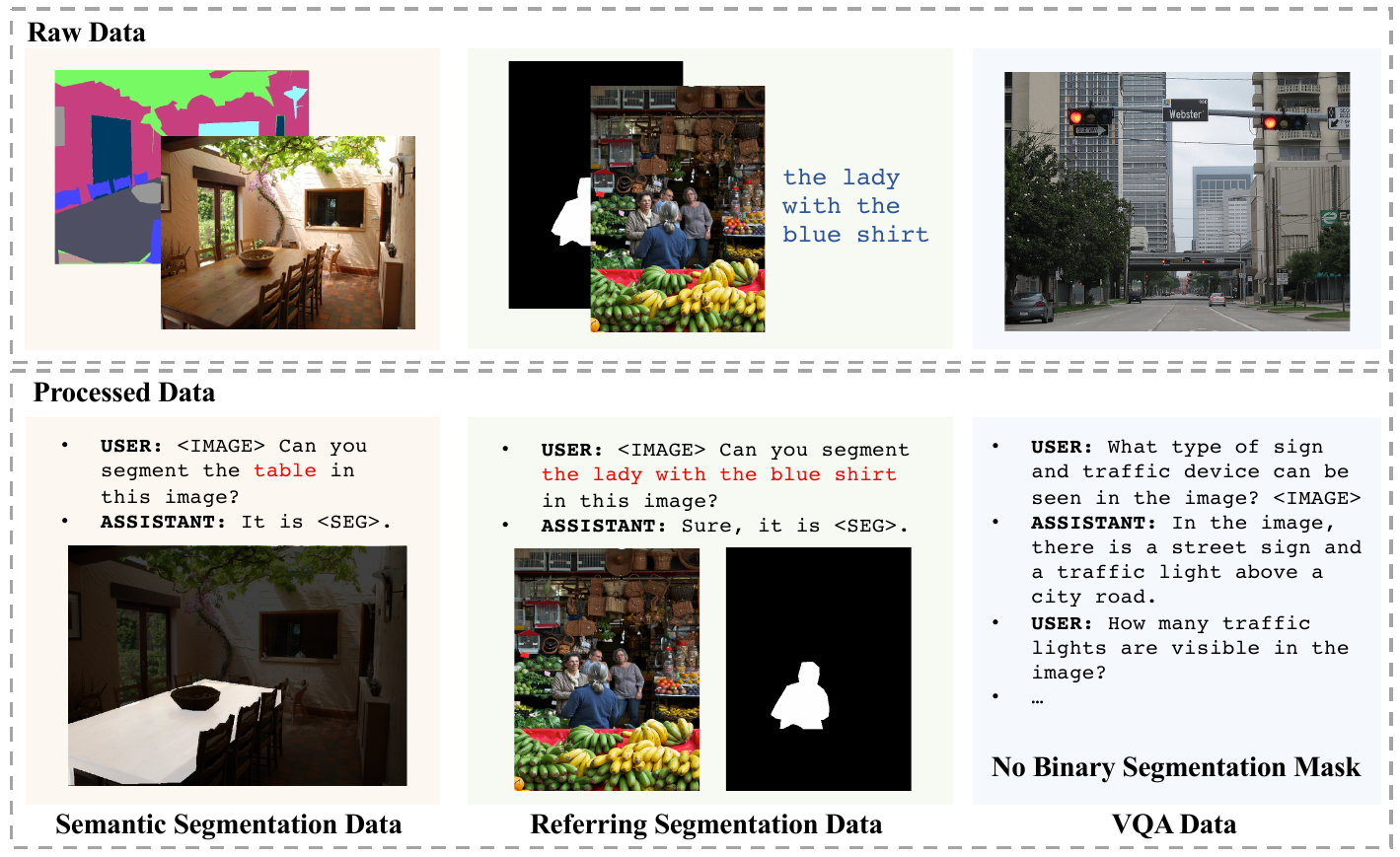}
\end{center}
\vspace{-0.6cm}
\caption{The illustration of training data formulation from different types of data, including semantic segmentation data, referring segmentation data, and visual question answering (VQA) data.}
\label{fig:data}
\vspace{-0.2cm}
\end{figure*}

\subsection{Training}
\label{method:training}

\paragraph{Training Data Formulation. } As illustrated in Fig.~\ref{fig:data}, our training data comprises mainly three parts, all of which are derived from widely-used public datasets. The details are as follows:

\begin{itemize}[leftmargin=0.3cm]
    \item \textit{Semantic Segmentation Dataset.} Semantic segmentation datasets typically consist of images and the corresponding multi-class labels. During training, we randomly choose several categories for each image. To generate data that matches the format of visual question answering, we employ a question-answer template like ``\texttt{\textbf{USER}}: \texttt{<IMAGE>} \texttt{\small Can you segment the} \texttt{\{class\_name\}} \texttt{\small in this image?} \texttt{\textbf{ASSISTANT}}: \texttt{\small It is} \texttt{<SEG>}.", where \texttt{\{class\_name\}} is the chosen category, and \texttt{<IMAGE>} denotes the placeholder for tokens of image patches. The corresponding binary segmentation mask is used as the ground truth to provide mask loss supervision. During training, we also use other templates to generate the QA data to ensure data diversity, as shown in the supplementary material. We adopt ADE20K, COCO-Stuff, and LVIS-PACO part segmentation datasets.
    \vspace{0.02cm}
    \item \textit{Vanilla Referring Segmentation Dataset.} Referring segmentation datasets provide an input image and an explicit short description of the target object. Thus, it is easy to convert them into question-answer pairs using a template like ``\texttt{\textbf{USER}}: \texttt{<IMAGE>}  \texttt{\small Can you segment} \texttt{\{description\}} \texttt{\small in this image?} \texttt{\textbf{ASSISTANT}}: \texttt{\small Sure, it is} \texttt{<SEG>}.", where \texttt{\{description\}} is the given explicit description. For this part, we adopt refCOCO, refCOCO+, refCOCOg, and refCLEF datasets.
    \vspace{0.02cm}
    \item \textit{Visual Question Answering Dataset.} To preserve the original Visual Question Answering (VQA) ability of the multimodal LLM, we also include the VQA dataset during training. We use LLaVA-Instruct-150k~\citep{liu2023visual} for LLaVA v1 and LLaVA-v1.5-mix665k for LLaVA v1.5~\cite{liu2023improved}.
\end{itemize}

Notably, the above datasets do not include any reasoning segmentation data sample. Instead, it only contains samples where the target objects are explicitly indicated in the query texts. Surprisingly, even without complex reasoning training data, LISA demonstrates impressive zero-shot ability on the \textit{ReasonSeg} benchmark, as shown in Table~\ref{table:reason_seg}. Moreover, we find that further performance boost could be yielded by finetuning the model on only 239 data samples that involve complex reasoning.

\begin{table*}[t]
    \centering
    \caption{Reasoning segmentation results among LISA (ours) and previous related works. `ft' denotes using 239 reasoning segmentation data samples to fine-tune the model. Unless otherwise specified, we use LLaVA v1~\cite{liu2023visual} as the base model. LLaVA1.5 denotes LLaVA v1.5~\cite{liu2023improved}.} 
    \vspace{-0.2cm}
    \label{table:reason_seg}   
    \tabcolsep=0.4cm
    {
        \begin{tabular}{ l | c c | c c | c c | c c }
            \toprule
            
            \multirow{3}*{Method} & \multicolumn{2}{c|}{val} & \multicolumn{6}{c}{test} \\ 
            
            \specialrule{0em}{0pt}{1pt}
            \cline{2-9}
            \specialrule{0em}{0pt}{1pt}
            
            
            ~ & \multicolumn{2}{c|}{overall} & \multicolumn{2}{c|}{short query} & \multicolumn{2}{c|}{long query} & \multicolumn{2}{c}{overall} \\

            \specialrule{0em}{0pt}{1pt}
            \cline{2-9}
            \specialrule{0em}{0pt}{1pt}
            
            ~ & gIoU & cIoU & gIoU & cIoU & gIoU & cIoU & gIoU & cIoU \\ 
            
            \specialrule{0em}{0pt}{1pt}
            \hline
            \specialrule{0em}{0pt}{1pt}

            OVSeg~\citep{liang2023open} & 28.5 & 18.6 & 18.0 & 15.5 & 28.7 & 22.5 & 26.1 & 20.8  \\



            GRES~\citep{liu2023gres} & 22.4 & 19.9 & 17.6 & 15.0 & 22.6 & 23.8 & 21.3 & 22.0 \\    %
            
            X-Decoder~\citep{zou2023generalized} & 22.6 & 17.9 & 20.4 & 11.6 & 22.2 & 17.5 & 21.7 & 16.3 \\

            SEEM~\citep{zou2023segment} & 25.5 & 21.2 & 20.1 & 11.5 & 25.6 & 20.8 & 24.3 & 18.7 \\
            
            Grounded-SAM~\citep{liu2023grounding} & 26.0 & 14.5 & 17.8 & 10.8 & 22.4 & 18.6 & 21.3 & 16.4 \\
            
            \specialrule{0em}{0pt}{1pt}
            \hline
            \specialrule{0em}{0pt}{1pt}
            
            LISA-7B & 44.4 & 46.0 & 37.6 & 34.4 & 36.6 & 34.7 & 36.8 & 34.1 \\

            LISA-7B (ft) & 52.9 & 54.0 & 40.6 & 40.6 & 49.4 & 51.0 & 47.3 & 48.4 \\
            
            \specialrule{0em}{0pt}{1pt}
            \hline
            \specialrule{0em}{0pt}{1pt}
            
            LISA-13B & 48.9 & 46.9 & 39.9 & 43.3 & 46.4 & 46.5 & 44.8 & 45.8 \\
            
            LISA-13B (ft) & 56.2 & 62.9 & 44.3 & 42.0 & 54.0 & 54.3 & 51.7 & 51.1 \\
            
            \specialrule{0em}{0pt}{1pt}
            \hline
            \specialrule{0em}{0pt}{1pt}
            
            LLaVA1.5-7B + OVSeg & 38.2 & 23.5 & 24.2 & 18.7 & 44.6 & 37.1 & 39.7 & 31.8 \\
            
            LISA-7B-LLaVA1.5 & 53.6 & 52.3 & 47.1 & 48.5 & 49.2 & 48.9 & 48.7 & 48.8 \\
            
            LISA-7B-LLaVA1.5 (ft) & 61.3 & 62.9 & 48.3 & 46.3 & 57.9 & 59.7 & 55.6 & 56.9 \\

            \specialrule{0em}{0pt}{1pt}
            \hline
            \specialrule{0em}{0pt}{1pt}
            
            LLaVA1.5-13B + OVSeg & 37.9 & 26.4 & 27.1 & 19.4 & 46.1 & 40.6 & 41.5 & 34.1 \\
            
            LISA-13B-LLaVA1.5 & 57.7 & 60.3 & 50.8 & 50.0 & 54.7 & 50.9 & 53.8 & 50.8 \\
            
            LISA-13B-LLaVA1.5 (ft) & \textbf{65.0} & \textbf{72.9} & \textbf{55.4} & \textbf{50.6} & \textbf{63.2} & \textbf{65.3} & \textbf{61.3} & \textbf{62.2} \\
            
            
            \bottomrule            
        \end{tabular}
    }
\vspace{-0.4cm}
\end{table*}

\vspace{-0.3cm}
\paragraph{Trainable Parameters. } 

To preserve the learned knowledge of the pre-trained multimodal LLM $\mathcal{F}$ (i.e., LLaVA in our experiments), we leverage LoRA~\citep{hu2021lora} to perform efficient fine-tuning, and completely freeze the vision backbone $\mathcal{F}_{enc}$. The decoder $\mathcal{F}_{dec}$ is fully fine-tuned. Additionally, the LLM token embeddings (\texttt{embed\_tokens}), the LLM head (\texttt{lm\_head}), and the projection layer $\gamma$ are also trainable. 

It is notable that the resulting model avoids the catastrophic forgetting of the original text generation capability and preserves the conversation ability, as verified in the supplementary material. The potential reasons are: we 1) employ LoRA fine-tuning to reduce the trainable parameters and 2) incorporate the VQA dataset during fine-tuning. 

\section{Experiment}


\vspace{-0.1cm}
\subsection{Experimental Setting}
\label{exp:setting}
\vspace{-0.1cm}
\paragraph{Network Architecture.}
Unless otherwise specified, we use LLaVA-7B-v1-1 or LLaVA-13B-v1-1~\cite{liu2023visual} as the base multimodal LLM $\mathcal{F}$, and adopt the ViT-H SAM~\citep{kirillov2023segment} backbone as the vision backbone $\mathcal{F}_{enc}$. The projection layer of $\gamma$ is an MLP with channels of [256, 4096, 4096]. 

\vspace{-0.3cm}
\paragraph{Implementation Details.}
We adopt 8 NVIDIA 24G 3090 GPUs for training. The training scripts are based on deepspeed~\citep{rasley2020deepspeed} engine. We use AdamW~\citep{loshchilov2017decoupled} optimizer with the learning rate and weight decay set to 0.0003 and 0, respectively. We also adopt WarmupDecayLR as the learning rate scheduler, where the warmup iterations are set to 100. The weights of the text generation loss $\lambda_{txt}$ and the mask loss $\lambda_{mask}$ are set to $1.0$ and $1.0$, respectively, and those of the bce loss $\lambda_{bce}$ and the dice loss $\lambda_{dice}$ are set to $2.0$ and $0.5$, respectively. Besides, the batch size per device is set to 2, and the gradient accumulation step is set to 10. During training, we select at most 3 categories for each image in semantic segmentation datasets.

\vspace{-0.3cm}
\paragraph{Datasets.}
As mentioned in Sec.~\ref{method:training}, our training data is mainly composed of three types of datasets: (1) For the semantic segmentation dataset, we use ADE20K~\citep{zhou2017scene} and COCO-Stuff~\citep{caesar2018coco}. Besides, to enhance the segmentation result for some part of an object, we also use part semantic segmentation datasets, including PACO-LVIS~\citep{ramanathan2023paco}, PartImageNet~\citep{he2022partimagenet}, and PASCAL-Part~\citep{chen2014detect}; (2) For the referring segmentation dataset, we use refCLEF, refCOCO, refCOCO+~\citep{kazemzadeh2014referitgame}, and refCOCOg~\citep{mao2016generation}; (3) For the visual question answering (VQA) dataset, we use the datasets of LLaVA-Instruct-150k for LLaVA v1~\citep{liu2023visual} and LLaVA-v1.5-mix665k for LLaVA v1.5~\cite{liu2023improved}. In order to avoid data leakage, we exclude the COCO samples whose images are present in the refCOCO(+/g) validation sets during training. Furthermore, we surprisingly find that by fine-tuning the model on only 239 ReasonSeg data samples, the model's performance can be further boosted.


\vspace{-0.3cm}
\paragraph{Evaluation Metrics.} We follow most previous works on referring segmentation~\citep{kazemzadeh2014referitgame, mao2016generation} to adopt two metrics: gIoU and cIoU. gIoU is defined by the average of all per-image Intersection-over-Unions (IoUs), while cIoU is defined by the cumulative intersection over the cumulative union. Since cIoU is highly biased toward large-area objects and it fluctuates too much, gIoU is preferred.

\begin{table*}[t]
    \centering
    \caption{Referring segmentation results (cIoU) among LISA (ours) and existing methods.}
    \label{table:refer_seg}   
    \vspace{-0.2cm}
    \tabcolsep=0.3cm
    {
        \begin{tabular}{ l | c c c | c c c | c c }
            \toprule
            
            \multirow{3}*{Method} & \multicolumn{3}{c|}{refCOCO} & \multicolumn{3}{c|}{refCOCO+}  & \multicolumn{2}{c}{refCOCOg} \\ 
            
            \specialrule{0em}{0pt}{1pt}
            \cline{2-9}
            \specialrule{0em}{0pt}{1pt}
            
            ~ & val & testA & testB & val & testA & testB & val(U) & test(U) \\ 
            
            
            
            \specialrule{0em}{0pt}{1pt}
            \hline
            \specialrule{0em}{0pt}{1pt}

            MCN~\citep{luo2020multi} & 62.4 & 64.2 & 59.7 & 50.6 & 55.0 & 44.7 & 49.2 & 49.4 \\

            VLT~\citep{ding2021vision} & 67.5 & 70.5 & 65.2 & 56.3 & 61.0 & 50.1 & 55.0 & 57.7 \\

            CRIS~\citep{wang2022cris} & 70.5 & 73.2 & 66.1 & 62.3 & 68.1 & 53.7 & 59.9 & 60.4 \\

            LAVT~\citep{yang2022lavt} & 72.7 & 75.8 & 68.8 & 62.1 & 68.4 & 55.1 & 61.2 & 62.1 \\
            
            ReLA~\citep{liu2023gres} & 73.8 & 76.5 & 70.2 & \textbf{66.0} & \textbf{71.0} & 57.7 & 65.0 & 66.0 \\
            
            X-Decoder~\citep{zou2023generalized} & - & - & - & - & - & - & 64.6 & -  \\

            SEEM~\citep{zou2023segment} & - & - & - & - & - & - & 65.7 & -    \\
            
            \specialrule{0em}{0pt}{1pt}
            \hline
            \specialrule{0em}{0pt}{1pt}
            
            LISA-7B & 74.1 & 76.5 & 71.1 & 62.4 & 67.4 & 56.5 & 66.4 & 68.5 \\
            
            LISA-7B (fine-tuned on ReferSeg) & \textbf{74.9} & \textbf{79.1} & \textbf{72.3} & 65.1 & 70.8 & \textbf{58.1} & \textbf{67.9} & \textbf{70.6} \\

            
            \bottomrule            
        \end{tabular}
    }
\vspace{-0.2cm}
\end{table*}

\subsection{Reasoning Segmentation Results}
\label{exp:reasonseg}

The reasoning segmentation results are shown in Table~\ref{table:reason_seg}. It is worth noting that existing works fail to handle the task, but our model can accomplish the task involving complex reasoning with more than $20\%$ gIoU performance boost. As mentioned before, the reasoning segmentation task is essentially different from the referring segmentation task in that it requires the model to possess \textit{reasoning ability} or access \textit{world knowledge}. Only by truly understanding the query, can the model do well in the task. The existing works have no proper way to understand an implicit query, but our model exploits multimodal LLMs to reach the goal. 

Notably, we also make a comparison with the vanilla two-stage method (LLaVA1.5 + OVSeg). Specifically, the two-stage method refers to first using a multimodal LLM (e.g., LLaVA v1.5) to generate a text output for the input query, and then adopting a referring or open-vocabulary segmentation model (e.g., OVSeg) to generate the segmentation mask. If the intermediate text output remains too long and exceeds the input token length limit of OVSeg, we use GPT-3.5 to further summarize. More details can be found in the supplementary material. The results in Table~\ref{table:reason_seg} show that our model outperforms the two-stage method significantly. We explain that the potential reasons are: 1) Our model is trained end-to-end, while the two-stage method is completely decoupled; 2) The two-stage method relies on text as an intermediary to transmit information, while our model utilizes the hidden embedding that is more expressive.

Another finding is that LISA-13B outperforms the 7B counterpart substantially, especially on the long-query scenarios, which indicates that the current performance bottleneck may still lie in understanding the query text, and a stronger multimodal LLM (e.g., LLaVA v1.5~\cite{liu2023improved}) leads to even better results.

\subsection{Vanilla Referring Segmentation Results}
\label{exp:referseg}

To show that our model is also competent in the vanilla referring segmentation task, we make a comparison with existing state-of-the-art methods in Table~\ref{table:refer_seg}. We evaluate the methods on refCOCO, refCOCO+, refCOCOg validation and testing sets. Our model achieves state-of-the-art results across various referring segmentation benchmarks.

\subsection{Ablation Study}
\label{exp:ablation}
In this section, we conduct an extensive ablation study to reveal the contribution of each component. Unless otherwise specified, we report the metrics of gIoU and cIoU of LISA-7B on the validation set.


\begin{table}[t]
    \footnotesize
    \centering
    \caption{Ablation study on the design choice of vision backbone. `ft' denotes finetuning on ReasonSeg training set.}
    \label{table:vision_backbone}
    \tabcolsep=0.6cm
    \begin{tabular}{c | c c}
        \toprule
        Vision Backbone & gIoU & cIoU \\
        \midrule
    
        Mask2Former-Swin-L & 42.4 & 38.8\\
    
        SAM (w/ LoRA) & 41.5 & 37.3\\
        
        SAM & 44.4 & 46.0 \\
        
        \specialrule{0em}{0pt}{1pt}
        \hline
        \specialrule{0em}{0pt}{1pt}
        
        
        Mask2Former-Swin-L (ft) & 50.7 & 52.3 \\
    
        
        SAM w/ LORA (ft) & 51.8 & 51.9 \\
        
        SAM (ft) & \textbf{52.9} & \textbf{54.0} \\
        \bottomrule
    \end{tabular}
\end{table}

\begin{figure*}[t]
\begin{center}
\includegraphics[width=0.95\linewidth]{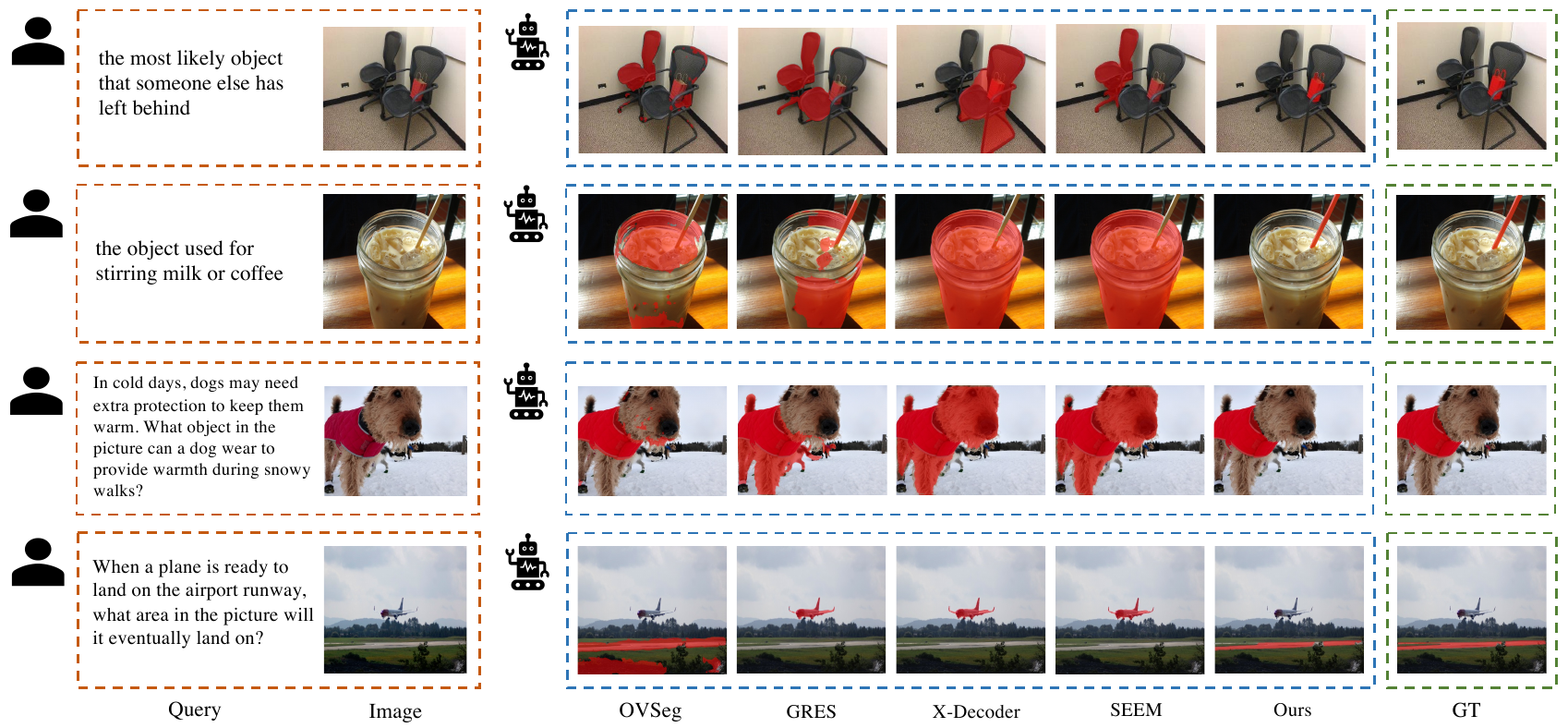}
\end{center}
\vspace{-0.7cm}
\caption{Visual comparison among LISA (ours) and existing related methods. More illustrations are given in the supplementary material.}
\label{fig:vis_comp}
\vspace{-0.5cm}
\end{figure*}

\begin{table}[t]
    \footnotesize
    \centering
    \caption{Ablation study on SAM pre-trained weight and rephrasing.}
    \label{table:ablation}
    \tabcolsep=0.2cm
    \begin{tabular}{c | c c | c c }
        \toprule
        Exp. ID & Pre-train\textsubscript{SAM} $\gamma$ & rephrasing & gIoU & cIoU \\
        \midrule
        1 &  & \Checkmark & 35.9 & 44.6 \\
        2 & \Checkmark &  & 50.7 & 51.1  \\
        3 & \Checkmark & \Checkmark & \textbf{52.9} & \textbf{54.0} \\
        \bottomrule
    \end{tabular}
\end{table}



    
    
    
    

    
    
\vspace{-0.2cm}
\paragraph{Design Choices of Vision Backbone.} We emphasize that vision backbones other than SAM are also applicable in our framework. In Table~\ref{table:vision_backbone}, we notice that SAM performs the best, potentially because of the massive high-quality data used in its pre-training phase. Further, we also find that with the Mask2Former backbone, our framework still achieves a decent performance on the reasoning segmentation task, significantly outperforming previous works such as X-Decoder~\citep{zou2023generalized}. This reveals the fact that the design choice of vision backbone is flexible and not limited to SAM.

\vspace{-0.2cm}
\paragraph{SAM LoRA Fintuning.} We also investigate the effectiveness of applying LoRA on the SAM backbone. In Table~\ref{table:vision_backbone}, we note that the performance of LoRA fine-tuned SAM backbone is inferior to that of the frozen one. A potential reason is that fine-tuning impairs the generalization ability of the original SAM model. 

\vspace{-0.2cm}
\paragraph{SAM Pre-trained Weight.} To demonstrate the contribution of SAM pre-trained weight, we make a comparison between Experiments 1 and 3 in Table~\ref{table:ablation}. Without being initialized with SAM pre-trained weight, the vision backbone is trained from scratch. This causes the performance to fall substantially behind that of the baseline model. 


\begin{table}[t]
\footnotesize
\centering
\caption{Ablation study on training data.}
\label{table:ablation_training_data}
\vspace{-0.2cm}
\tabcolsep=0.04cm
{
\begin{tabular}{c | c c c | c | c | c | c c}
    \toprule
    \multirow{2}*{ID} & \multicolumn{3}{c|}{SemanticSeg} & \multirow{2}*{ReferSeg} & \multirow{2}*{VQA} & \multirow{2}*{ReasonSeg} & \multirow{2}*{gIoU} & \multirow{2}*{cIoU} \\
    
    \specialrule{0em}{0pt}{1pt}
    \cline{2-4}
    \specialrule{0em}{0pt}{1pt}
    
    ~ & ADE20K & COCO-Stuff & PartSeg & ~ & ~ & ~ & ~ & ~ \\
    
    \specialrule{0em}{0pt}{1pt}
    \hline
    \specialrule{0em}{0pt}{1pt}

    1 &  & \Checkmark & \Checkmark & \Checkmark & \Checkmark & \Checkmark & 48.9 & 53.5  \\
    2 &  \Checkmark &  & \Checkmark & \Checkmark & \Checkmark & \Checkmark & 48.5 & 50.8  \\
    3 &  \Checkmark & \Checkmark & & \Checkmark & \Checkmark & \Checkmark & 46.7 & 50.9\\
    4 &   &  & \Checkmark & \Checkmark & \Checkmark & \Checkmark & 46.6 & 46.7  \\
    5 &   &  &  & \Checkmark & \Checkmark & \Checkmark & 30.4 & 20.4  \\
      
    \specialrule{0em}{0pt}{1pt}
    \hline
    \specialrule{0em}{0pt}{1pt}
    
    6 & \Checkmark & \Checkmark & \Checkmark &  & \Checkmark & \Checkmark & 47.7 & 51.1 \\
    
    \specialrule{0em}{0pt}{1pt}
    \hline
    \specialrule{0em}{0pt}{1pt}
    
    7 & \Checkmark & \Checkmark & \Checkmark & \Checkmark & \Checkmark & & 44.4 & 46.0  \\
    
    \specialrule{0em}{0pt}{1pt}
    \hline
    \specialrule{0em}{0pt}{1pt}
    
    8 & \Checkmark & \Checkmark & \Checkmark & \Checkmark & \Checkmark & \Checkmark & \textbf{52.9} & \textbf{54.0}  \\
    
    \bottomrule
\end{tabular}
}
\vspace{-0.2cm}
\end{table}


\vspace{-0.2cm}
\paragraph{Instruction Rephrasing by GPT-3.5.} When fine-tuning the model on the reasoning segmentation data samples, we rephrase the text instruction by GPT-3.5 (the details are shown in the supplementary material), and randomly choose one. The comparison between Experiments 2 and 3 in Table~\ref{table:ablation} shows that the performance is increased by 2.2\% gIoU and 2.9\% cIoU. This result verifies the effectiveness of such data augmentation.

\vspace{-0.2cm}
\paragraph{Contribution of All Types of Training Data.}
In Table~\ref{table:ablation_training_data}, we show the contribution of each type of data to the performance. We find that in Exp. 5, we do not use any semantic segmentation dataset, and the performance drops a lot. We conjecture that semantic segmentation datasets provide a large amount of ground-truth binary masks for training, since a multi-class label can induce multiple binary masks. 

\begin{table}[t]
\footnotesize
\centering
\caption{Results on the ReasonSeg test set.}
\label{table:ablation_reasonseg_data}
\vspace{-0.2cm}
\tabcolsep=0.3cm
{
\begin{tabular}{l | c | c c}
    \toprule

    Training splits & \# data samples & gIoU & cIoU \\

    \specialrule{0em}{0pt}{1pt}
    \hline
    \specialrule{0em}{0pt}{1pt}

    train & 239 & 51.7 & 51.1 \\
    
    \specialrule{0em}{0pt}{1pt}
    \hline
    \specialrule{0em}{0pt}{1pt}
    
    train + val & 439 & \textbf{54.0} & \textbf{54.9} \\
    
    \bottomrule
\end{tabular}
}
\vspace{-0.4cm}
\end{table}

We also notice that adding more reasoning segmentation data samples during training leads to better results. In Table~\ref{table:ablation_reasonseg_data}, we also add the ReasonSeg val set (200 data samples) during fine-tuning, and it yields better performance in both gIoU and cIoU metrics. This indicates that more reasoning segmentation training samples are beneficial at this moment.

\subsection{Qualitative Results}
\label{exp:qualitative}

As depicted in Fig.~\ref{fig:vis_comp}, we provide a visual comparison with existing related works, including the model for open-vocabulary semantic segmentation (OVSeg), referring segmentation (GRES), and the generalist models for segmentation (X-Decoder and SEEM). These models fail to handle the displayed cases with various errors, while our approach produces accurate and high-quality segmentation results. More illustrations are given in the supplementary material.

\section{Conclusion}


In this work, we have proposed a new segmentation task—\textit{reasoning segmentation}. Also, we have introduced an evaluation benchmark \textit{ReasonSeg}, which comprises over one thousand data samples. Finally, we have presented our model --- LISA. It injects segmentation capabilities into current multimodal LLMs and performs surprisingly effectively on the reasoning segmentation task. We hope our work can shed new light on the direction of combining LLMs and vision tasks in the future.

\vspace{-0.2cm}
\section*{Acknowledgements}
\vspace{-0.1cm}
This work is supported in part by the Research Grants Council under the Areas of Excellence scheme grant AoE/E-601/22-R and the Shenzhen Science and Technology Program under No. KQTD20210811090149095.

{
    \small
    \bibliographystyle{ieeenat_fullname}
    \bibliography{main}
}


\end{document}